\documentclass[runningheads]{llncs}
\usepackage{graphicx}
\usepackage{caption}
\usepackage{subcaption}
\usepackage[export]{adjustbox}
\usepackage{verbatim}
\usepackage{bbm}
\usepackage{multirow}
\usepackage{amsmath}
\usepackage[symbol]{footmisc}
\begin{document}
\title{mulEEG: A Multi-View Representation Learning on EEG Signals}
%
%

\newcommand*\samethanks[1][\value{footnote}]{\footnotemark[#1]}
\renewcommand{\thefootnote}{\fnsymbol{footnote}}

\author{Vamsi Kumar \thanks{Equal Contribution}  \inst{1}   \and
Likith Reddy \samethanks \inst{1}  \and
Shivam Kumar Sharma\inst{1} \and
Kamalakar Dadi \inst{1} \and
Chiranjeevi Yarra \inst{1} \and
Bapi Raju \inst{1} \and
Srijithesh Rajendran \inst{2}}
 \authorrunning{Vamsi Kumar and Likith Reddy et al.}
%
\institute{International Institute of Information Technology, Hyderabad, India \and
National Institute of Mental Health and Neuro Sciences, Bangalore, India.\\
\email{likith.reddy@ihub-data.iiit.ac.in}\\
\url{http://ihub-data.iiit.ac.in}}

\maketitle              

\begin{abstract}
Modeling effective representations using multiple views that
positively influence each other is challenging, and the existing
methods perform poorly on Electroencephalogram (EEG) signals for sleep-staging tasks. In this paper, we propose a novel multi-view self-supervised method (mulEEG) for unsupervised EEG representation learning. Our method attempts to effectively utilize the complementary information available in multiple views to learn better representations. We introduce diverse loss that further encourages complementary information across multiple views. Our method with no access to labels, beats the supervised training  while outperforming multi-view baseline methods on transfer learning experiments carried out on sleep-staging tasks. We posit that our method was able to learn better representations by using complementary multi-views.
\footnote[2]{Code Available at : https://github.com/likith012/mulEEG}

\keywords{Multi-View Learning  \and Self-Supervised \and Sleep Staging}

\end{abstract}

\section{Introduction}
\indent \par{} Sleep is an important part of daily routine. Getting the right quantity and quality of sleep at the right time is essential for one’s well-being. As identification of sleep stages plays a crucial role in determining the quality of sleep, it can also help identify sleep-related or other mental disorders like Obstructive Sleep Apnea, depression, schizophrenia, and dementia \cite{bianchi2010obstructive,freeman2020sleep,gottesmann2007neurobiological}.

With the advent of the large volume of devices that can monitor Electroencephalogram(EEG) signals, the amount of data piling up is enormous. Many studies have proposed exploiting such EEG data for automated sleep stage classification using supervised methods \cite{eldele2021attention,sors2018convolutional,supratak2017deepsleepnet,zhu2014analysis}. However, such methods rely on massive amounts of annotated data. Annotating EEG data is costly and time-consuming for a physician. The inter-rater agreement on annotations for different sleep stages is low \cite{younes2018reliability}, making annotations unreliable. A model trained on EEG data annotated by one expert will be biased towards their annotations and may not generalize well. Recently, self-supervised learning (SSL) has been used to learn effective representations from unlabeled data. Various self-supervised methods have been proposed on natural image datasets that rely on contrastive loss \cite{chen2020simple,he2020momentum,oord2018representation}, clustering \cite{caron2018deep,caron2020unsupervised}, or distillation \cite{grill2020bootstrap} utilizing massive amounts of unlabeled data. Limited work has been done on self-supervised learning for EEG signals, \cite{banville2021uncovering} proposed SSL tasks that apply relative positioning and contrastive predictive coding on time-series EEG signals. \cite{eldele2021time} used contrastive loss while addressing the temporal dependencies in time-series more effectively. \cite{mohsenvand2020contrastive} extended the simCLR \cite{chen2020simple} framework to time-series EEG signals while introducing new augmentations. \cite{yang2021self} reduced the impact of negative sampling usually caused due to contrastive loss, on spectrograms from EEG signals.

Multi-view SSL jointly trains all the views influencing each other in a self-supervised way. On natural image datasets, \cite{tian2020contrastive} captured the shared information across multiple views by maximizing mutual information between them. \cite{han2020self} focused on learning representations for videos in a self-supervised way by using complementary information from two different views (RGB and optical flow). \cite{yuan2021multimodal} learned visual representations by using multi-modal data with a combination of inter- and intra-modal similarity preservation objectives. The same multi-view methods cannot be applied directly to EEG signals as augmentation strategies are different from images. Moreover, extracting multiple views for EEG signals differs from that on natural images. To the best of our knowledge, on physiological signals no work has been done on multi-view SSL, which aims to learn effective representations by training multiple views jointly.

Our objective is to learn effective representations from multiple views by training them jointly in a self-supervised way. We aim to utilize the complementary information in multiple views to influence each other positively during training. The main contributions of this work are as follows: 1) We design an EEG augmentation strategy for multi-view SSL. 2) We illustrate that existing multi-view self-supervised methods perform poorly and inconsistently on EEG data. 3) We propose a novel multi-view SSL method that effectively utilizes the complementary information available in multiple views to learn better representations. 4) We introduce an additional diverse loss function that further encourages the complementary information across multiple views.

\section{Methods}

\indent\par{\textbf{Data Augmentation:}}
The contrastive learning methods are heavily influenced by the stochastic data augmentations used \cite{grill2020bootstrap}, and it is important to get the right match of augmentations. We use a different family of augmentations as it produces strong variations between the two augmented views and is known to work better when a shared encoder is used \cite{chen2020simple}. Similar augmentation approaches are used in \cite{eldele2021time,mohsenvand2020contrastive} for EEG signals. We use jittering, where random uniform noise is added to the EEG signal depending on its peak-to-peak values, along with masking, where signals are masked randomly as one family of augmentations $T_1$. Flipping, where the EEG signal is horizontally flipped randomly, and scaling, where EEG signal is scaled with Gaussian noise are used sequentially as the second family of augmentations $T_2$.

\begin{figure}[t]
     \centering
     \begin{subfigure}[b]{0.3\textwidth}
         \centering
         \includegraphics[width=\textwidth]{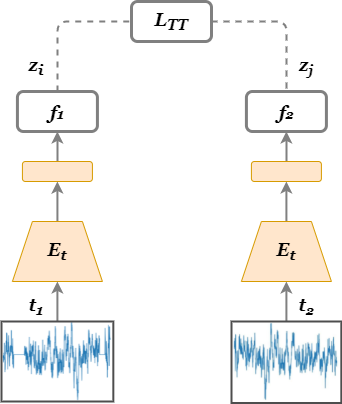}
         \caption{Single-View}
         \label{fig: single_view}
     \end{subfigure}
     \hfill
     \begin{subfigure}[b]{0.6\textwidth}
         \centering
         \includegraphics[width=\textwidth]{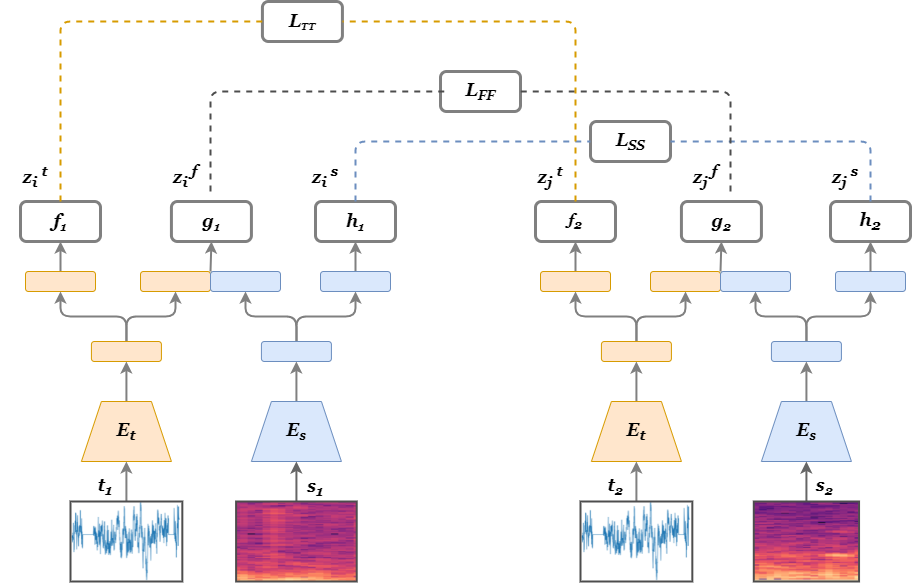}
         \caption{mulEEG}
         \label{fig: our_model}
     \end{subfigure}
        \caption{Architecture overview of Single-View and our proposed method mulEEG}
        \label{fig: model_architecture}
\end{figure}

\par{\textbf{Problem Formulation:}}
The input is a continuous recording of an EEG signal from a subject and is segmented into non-overlapping 30 seconds EEG signal called an epoch. Each epoch is categorized into five sleep stages: Wake, REM, N1, N2, N3 according to American Academy of Sleep Medicine \cite{berry2017aasm} guidelines. And
each subject’s EEG recording epochs belongs to one of the pretext/train/test
groups where the pretext group contains a relatively large number of unlabeled subjects compared to the less number of labeled subjects in train and test groups. Throughout our study, time-series and spectrogram are used as multiple views of the same EEG signal. For each EEG epoch, we obtain two time-series augmentations denoted as $t_1 \sim T_1$ and $t_2 \sim T_2$. The time-series augmentations are converted into their respective spectrograms $s_1$, $s_2$. The augmentations are passed into their respective encoders, which are time-series encoder, $E_t$, and spectrogram encoder, $E_s$, to extract their high dimensional latent representations. A separate projection head is used for the time-series and spectrogram encoder, which takes input as these high dimensional latent representations and maps them to the space where contrastive loss is applied on $\mathbf{z_i}$, $\mathbf{z_j}$. Applying the contrastive loss directly on the projection head outputs gives better results \cite{chen2020simple}. We use a variant of contrastive loss called NT-Xent \cite{chen2020simple,oord2018representation} which maximizes the similarity between two augmented views while minimizing its similarity with other samples. Here, $N$ is the batch size, $\tau$ is the temperature parameter, and a cosine similarity is used in the contrastive loss function given in \eqref{eq:2}. The function ${\mathbbm{1}}_{[k\ne{i}]} \epsilon \{0,1\}$ evaluates to 1 if $k\ne{i}$, otherwise gives 0.

\begin{align}\ell(i,j) = -\log \frac{exp(cosine(\mathbf{z_i}, \mathbf{z_j})/\tau)))}{\sum_{k=1}^{2N} {\mathbbm{1}}_{[k\ne{i}]}  exp(cosine(\mathbf{z_i},\mathbf{z_k})/\tau))}, \label{eq:1}
\end{align}
\begin{align}
L(\mathbf{z_i},\mathbf{z_j}) = \frac{1}{2N}{\sum_{k=1}^{N}\ell(2k-1, 2k)+\ell(2k, 2k-1))} , \label{eq:2}
\end{align}

\par{\textbf{Single-View and Multi-View:}}
In single-view learning Fig.1(a), we learn the representations for time-series, $E_t$,  and spectrogram encoder, $E_s$, by training them independently using unlabeled pretext group. Inspired from \cite{chen2020simple}, the encoder is shared and is trained on a pair of augmentations. For pre-training time series encoder, $E_t$, the time-series augmentations $t_1$, $t_2$ are passed to $E_t$ and then its output to the time-series projection heads $f_1$, $f_2$ to obtain $\mathbf{z_i}$ and $\mathbf{z_j}$. The contrastive loss $L_{TT}$ in \eqref{eq:2} is applied on  $\mathbf{z_i}$,  $\mathbf{z_j}$ and similarly, the spectrogram encoder, $E_s$, is pre-trained.

In a multi-view setup, we learn the representations of time-series and spectrogram encoder by pre-training them jointly, thereby influencing each other. We present two baseline multi-view methods: 1) \textbf{CMC}: Inspired from Contrastive Multiview Coding \cite{tian2020contrastive}, we jointly train both the encoders by sending $t_1$ to $E_t$ and $s_2$ to $E_s$, and the outputs from encoders are sent to their projection heads $f_1$, $f_2$ to obtain $\mathbf{z_i}$ and $\mathbf{z_j}$. Compared to \cite{tian2020contrastive} we apply augmentations in our method and use a projection head. It brings both the encoders latent features $\mathbf{z_i}$, $\mathbf{z_j}$ into the same feature space and compares them by aiming for information maximization between the views.  2) \textbf{Simple Fusion}: We experiment with a setup where we jointly train both the encoders by considering the concatenation of features obtained from $E_t$ and $E_s$. The augmentations $t_1$, $s_1$ are passed through $E_t$, $E_s$, and the concatenated outputs are passed through a projection head $f_1$ to obtain $\mathbf{z_i}$. Similarly, for augmented views  $t_2$, $s_2$, we extract $\mathbf{z_j}$ and contrastive loss is applied on concatenated feature space. \par{\textbf{mulEEG:}}
In a supervised setup, when both views (time-series \& spectrogram) are trained jointly, predicting on concatenated outputs does not always guarantee a better performance when compared to a setup where a single-view is trained individually for sleep-stage classification \cite{phan2021xsleepnet}. It is observed that such a setup behaves differently under different datasets and modalities. This behavior is somewhat surprising and is not well understood. Also, it is observed on multiple datasets with multiple modality combinations that, along with encoders, time-series and spectrogram views are favorable for identifying different sleep stages. But this favourability on sleep stages is not constant and changes from dataset to dataset. From \cite{phan2021xsleepnet}, we cannot say that time-series or spectrogram views are better than others, but instead, one should aim to use both views to complement each other when trained jointly, trying to optimize the task at hand. 

In our method shown in Fig.1(b), we jointly train the encoders $E_t$, $E_s$ such
that the complementary information of both views is utilized optimally and positively influences each other. The time, spectrogram, and concatenation of time and spectrogram features are used in our method along with the shared encoders $E_t$, $E_s$. The augmented views of time-series and spectrogram $t_1$, $s_1$ are passed through their respective encoders. Apart from obtaining the time and spectrogram features, we also concatenate time and spectrogram features to be further given as input to three different projection heads $f_1$, $g_1$, $h_1$. Similarly for other augmented views $t_2$, $s_2$ output from the encoders is sent to $f_2$, $g_2$, $h_2$. It uses three contrastive losses, each loss working on the time-series feature space $L_{TT}$, spectrogram feature space $L_{SS}$ and concatenated features space $L_{FF}$. Such a setup introduces flexibility to optimize between the time, spectrogram, and concatenated features during the self-supervised training.

\indent We introduce an additional loss called Diverse Loss $L_D$ given in  \eqref{eq:4} that further encourages the complementary information across time-series and spectrogram views. Due to contrastive loss on concatenated features, the time-series and spectrogram representations can tend to maximize the mutual information between them. Such a process can ignore the complementary information inherent to the views. Different from the contrastive loss used above, here the contrastive loss is applied on features obtained from only time-series and spectrogram projection heads $z_k = [\mathbf{z_{i}^t},\mathbf{z_{j}^t},\mathbf{z_{i}^s},\mathbf{z_{j}^s}] \; \forall i=j$ on a single sample $k$ instead on entire batch $N$. The contrastive loss here tries to pull time-series features closer while pushing away spectrogram features from time-series features for a single sample. Similarly, spectrogram features are pushed closer while pushing away time-series features for a sample. This allows the representations learned by both time-series and spectrogram views to have diverse information from each other for a single sample. The total loss given in \eqref{eq:5}, is a combination of contrastve loss on time-series, spectrogram, and concatenated features that in turn combined with
diverse loss. Our method can be extended to use along with recent SSL strategies, using moving average encoder \cite{he2020momentum,grill2020bootstrap}, negative sampling strategies \cite{yang2021self}, etc.

\begin{align}
\ell_d(z_k,a,b) = -\log\frac{exp(cosine(z_k[a],z_k[b])/\tau_d)}{\sum_{i=1}^{4} {\mathbbm{1}}_{[i\ne{a}]} exp(cosine(z_k[a],z_k[i])/\tau_d))}, \label{eq:3}
\end{align}

\begin{align}
L_{D} = \frac{1}{4N} \sum_{k=1}^{N} \ell_d(z_k,1,2) + \ell_d(z_k,2,1) + \ell_d(z_k,3,4) + \ell_d(z_k,4,3),\label{eq:4}
\end{align}

\begin{align}
L_{tot} = \lambda_1(L_{TT}+L_{FF}+L_{SS})+\lambda_2L_D \label{eq:5}
\end{align}

\section{Experiments}
\indent\par{\textbf{Datasets:}} To evaluate our proposed method, we consider two popular publicly available sleep-staging datasets: 1) \textbf{SleepEDF:} The sleep-EDF \cite{physionet} database contains a collection of 78 subjects to understand the age effects on sleep in healthy Caucasians. Each subject contains two full night recordings, with a few subjects having only one, with a total of 153 full night recordings. We randomly shuffle subjects and select 58 subjects as a pretext, 10 as train, and 10 as test groups. We use a single-channel EEG (Fpz-Cz) sampled at 100 Hz. 2) \textbf{SHHS:} The Sleep Heart Health Study (SHHS) \cite{zhang2018national} is a multi-center cohort study comprising 6,441 subjects with a single full night recording for each subject. We have selected 326 subjects from the total subjects based on the criteria that the selected subjects are close to having a regular sleep cycle similar to \cite{eldele2021attention}. A single-channel EEG (C4-A1) at a sampling rate of 125 Hz is used, we convert the signals to 100 Hz to keep the experiments consistent with the sleepEDF dataset. We then randomly shuffle the subjects and use a data split of 264 (pretext), 31 (train), and 31 (test).
\par{\textbf{Implementation Details:}} In all our experiments, we use ResNet-50 \cite{he2016deep} with 1D-convolutions as a time series encoder with 0.6 million parameters. For the first 1D-convolution layer a kernel size of 71 is used and for the rest, a kernel size of 25 is used which outputs a 256-dimensional feature vector. For the spectrogram encoder, \cite{yang2021self} is used which takes in a spectrogram as input and outputs a 256-dimensional feature vector. For converting the augmented time-series EEG signal into a spectrogram, Short Time Fourier Transform is used with the number of FFT points set to 256 and hop length to 64. A non-linear two-layer MLP projection head is used as it performs better compared to a linear projection head \cite{chen2020simple}, and we include Batch Norm as it improves the performance drastically \cite{chen2021exploring}. The projection head architecture is Linear $>$ BatchNorm $>$ ReLU $>$ Linear outputting a 128-dimensional vector to be used in the contrastive loss space. During the self-supervised pre-training, Adam optimizer with an initial  learning rate of 3e-4 with $\beta_1 = 0.9$, $\beta_2 = 0.99$, and a weight decay of 3e-5 is used. It is trained for a total of 140 epochs with a batch size of 256 and a learning scheduler is used which reduces the learning rate by 1/5th with a patience of 5. For training the encoders in a supervised way, similar optimization parameters are used, but it is trained for a total of 300 epochs, batch size of 256 with learning scheduler reducing the learning rate by 1/5th with patience of 10. The temperature $\tau$ used for our model losses $L_{SS}$, $L_{TT}$, and $L_{FF}$ is 1, 1, 1, respectively. For the diverse loss $L_D$ the temperature $\tau_d$ used is 10 with $\lambda_1=1$ and $\lambda_2=1$ in the final loss.
\par{\textbf{Evaluation:}} The encoders are pre-trained using the unlabeled pretext group and evaluated on standard linear benchmarking evaluation protocol \cite{chen2020simple,chen2021exploring,he2020momentum,grill2020bootstrap}. In linear evaluation, a linear classifier is attached on top of the frozen pre-trained encoder and only the linear classifier is trained on the train group, and the metrics are evaluated on the test group. We evaluate the experiments on datasets in two ways: 1) \textbf{Within Dataset}: One is to obtain the pretrain/train/test groups within the dataset itself as done in \cite{eldele2021time,yang2021self,mohsenvand2020contrastive} and perform a 5-fold evaluation on train and test groups. 2) \textbf{Transfer Learning}: Another is \cite{chen2020simple,he2020momentum}, where we pre-train on larger dataset SHHS with 264 subjects as a pretext group and evaluate on another dataset, SleepEDF with 58 subjects(pretext group in within dataset) as train and 20 subjects(we add both train and test group in within dataset) as a test group. We use the metrics Cohen's kappa ($\kappa$), accuracy, and macro-averaged F1 (MF1) score to evaluate the performance. Macro-averaged F1 score considers the class-wise performance and performs better when each class performs better compared to other metrics. We also compare our method with supervised baselines: 1) \textbf{Supervised}: The encoder is trained in a supervised way on the pretext group and evaluated on train/test groups. 2) \textbf{Randomly Initialized}: The encoder is randomly initialized and isn’t trained on the pretext group. It is then further evaluated on train/test groups. These two baselines are commonly treated as lower and upper bounds to measure the quality of the learned representations using self-supervised learning \cite{arandjelovic2017look,korbar2018cooperative}.
\begin{table}[b]

\caption{Linear evaluation performance on our proposed method and baseline models for within dataset and transfer learning}

\centering
\begin{tabular}{|l|c c c|c c c|c c c|}
\hline
\multirow{3}{4em}{\textbf{Method}} & \multicolumn{6}{c|}{\textbf{Within Dataset}} & \multicolumn{3}{c|}{\textbf{Transfer Learning}}\\
\cline{2-10}
& \multicolumn{3}{c|}{SleepEDF} & \multicolumn{3}{c|}{SHHS} & \multicolumn{3}{c|}{SHHS $>$ SleepEDF}\\
\cline{2-10}
 & Acc & $\kappa$ & MF1 & Acc & $\kappa$ & MF1 & Acc & $\kappa$ & MF1 \\
\hline
Single-View   &77.58&0.6773&66.74&79.56&0.7145&64.71&76.73&0.6669&66.42\\
Simple Fusion &77.01&0.6683&65.71&79.39&0.7122&64.56&76.75&0.6658&65.78\\
CMC &74.21&0.6300&62.81&80.29&0.7215&65.93&75.84&0.6520&64.4\\
Ours    &77.84&0.6806&67.04&80.13&0.7223&65.57&78.18&0.6869&67.88\\
Ours + diverse loss &\textbf{78.06}&\textbf{0.6850}&\textbf{67.82}&\textbf{81.21}&\textbf{0.7366}&\textbf{66.58}&\textbf{78.54}&\textbf{0.6914}&\textbf{68.10}\\
\hline
Randomly Initialized            &40.52&0.1189&17.04&44.75&0.0894&19.39&38.68&0.1032&16.54\\
    Supervised &\textbf{79.08}&\textbf{0.7014}&\textbf{69.78}&\textbf{82.62}&\textbf{0.7569}&\textbf{71.41}&{77.88}&{0.6838}&{67.84}\\
\hline
\end{tabular}
\end{table}
\label{lin_eval_results}

\section{Results}
\indent\par{\textbf{Within Dataset:}} After jointly pre-training both the encoders using the multi-view setup, the representations learned by the time-series and spectrogram encoder can be evaluated individually or on concatenated features. In our experiments, we only evaluate the time-series encoder similar to \cite{tian2020contrastive,yuan2021multimodal}. The performance metrics for linear evaluation protocol are shown in Table 1, comparing our method with others. All the self-supervised methods outperform the Randomly Initialized model by a drastic margin, indicating that the encoders learn useful representations from unlabeled data using self-supervised methods. Simple Fusion tends to perform worse than the single-view method on both the datasets even though the encoders are pre-trained jointly, implying the spectrogram encoder wasn’t able to influence the time-series encoder positively. But the performance of CMC gives inconsistent results and performs much worse on sleepEDF but performs better on SHHS dataset compared to the single-view. Our method without the diverse loss performs consistently better on both datasets when compared to the single-view and CMC methods. But our model with diverse loss seems to perform much better consistently compared to all the self-supervised methods for both the datasets by a good margin. Compared to the single-view method, we observe an improvement of 3.1\% for $\kappa$ and 2.9\% for MF1 on SHHS dataset. This shows in our method when both encoders are pre-trained jointly, spectrogram encoder was able to influence the time-series encoder positively. The supervised model seems to outperform all methods on linear evaluation.\par{\textbf{Transfer Learning:}} The results for transfer learning are shown in Table 1, on linear evaluation. Similarly, we can observe all the self-supervised methods outperforming the Randomly Initialized model by a large margin. The single-view performs better than multi-view methods: Simple Fusion and CMC. Our model without the diverse loss performs better than the supervised baseline. When the diverse loss is included in our method, even better performance can be observed. When compared to the supervised baseline which has access to all the labels in pretext group, we can see both our methods with and without diverse loss, which were pre-trained with no labels, performing better than the supervised baseline. Compared to the supervised baseline, we see an improvement of 1.1\% on $\kappa$ and 0.85\% on accuracy. This is because supervised learning is biased to annotations generated by a physician on the SHHS dataset and couldn't generalize well to the SleepEDF dataset. This shows that the representations learned by the time-series encoder using our method are better than those learned using supervised learning on a sleep-staging task.
\begin{figure}[t]
     \centering
     \begin{subfigure}[b]{0.32\textwidth}
         \centering
         \includegraphics[width=\textwidth]{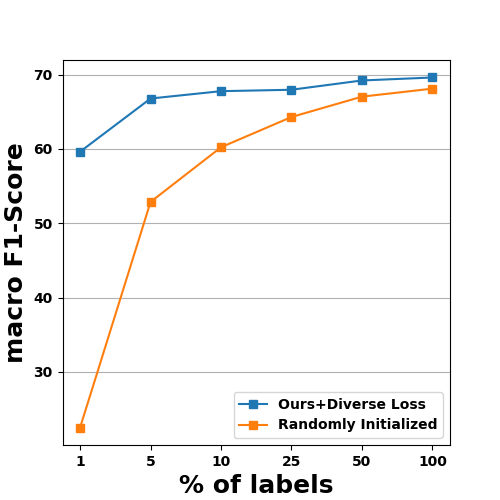}
         \caption{SleepEDF}
         \label{fig: sleepEDF_sem_sup}
     \end{subfigure}
     \hfill
     \begin{subfigure}[b]{0.32\textwidth}
         \centering
         \includegraphics[width=\textwidth]{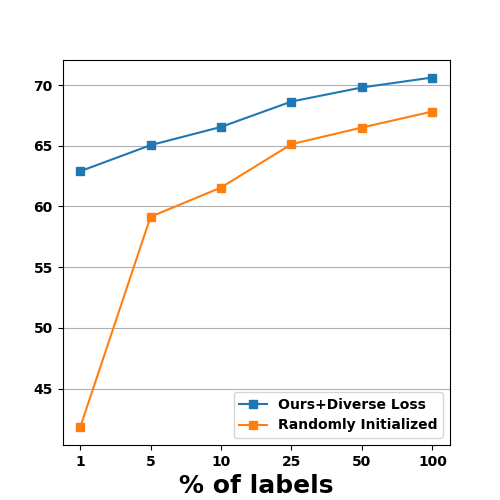}
         \caption{SHHS}
         \label{fig: shhs_sem_sup}
     \end{subfigure}
     \hfill
     \begin{subfigure}[b]{0.32\textwidth}
         \centering
         \includegraphics[width=\textwidth]{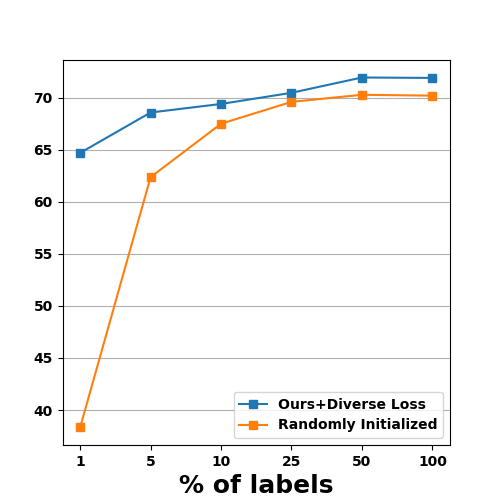}
         \caption{SHHS $>$ SleepEDF}
         \label{fig: d2d_sem_sup}
     \end{subfigure}
     \hfill
        \caption{Semi-supervised performance comparison between our method and randomly initialized on MF1 for within dataset and transfer learning }

        \label{semi_supervised_table}
\end{figure}\par{\textbf{Semi-supervised:}} To evaluate the effectiveness of our method under semi-supervised settings, we finetune with different amounts of training data and compare the results of time-series encoder pre-trained using our method with the supervised model initialized randomly. We randomly select 1\%, 5\%, 10\%, 25\%, 50\% and 100\% of samples from train group and show the results in Fig.\ref{semi_supervised_table}. When only 1\% of labeled data is given, the supervised method performs poorly while our method performs better by a significant margin for both within dataset and transfer learning experiments. This shows the effectiveness of our method when a limited amount of labeled data is available. With only 5\% of labeled data, our method is almost able to match the performance of randomly initialized supervised model when trained with 100\% labeled data. As the amount of labeled data increases, the gap between both the methods tends to decrease, but even when we use 100\% of labeled data, our method still performs better consistently.

\section{Conclusion}\indent\par This study proposed a novel multi-view SSL method called mulEEG for unsupervised EEG representation learning. Our method bootstraps the complementary information available in multiple views to learn better representations. While our method is general, specific application on sleep-stage classification is demonstrated. We show that on linear evaluation, our method was able to beat the supervised training on transfer learning and shows high efficiency on few-labeled scenarios.

\section*{Appendix: }

{\textbf{t-SNE Visualization:}} To understand the structures captured by our method,
we obtain the 256-dimensional output features from the time-series encoder and
analyze them using sleep-staging annotations. To visualize, we use t-SNE performed on the sleepEDF dataset, where 1000 samples are randomly selected from each class. In Fig.1 clear clusters are observed for each sleep stage. Apart from clusters, the distribution plot shows that the clusters are arranged
sequentially from left to right, starting at W, then going through N1, N2 and
N3 sequentially. This trajectory represents the sleep stage progression observed
clinically. REM is observed to be overlapping with the N1 sleep stage, which is in line with previous observations on the structure of the sleep-wakefulness continuum .
\begin{figure}[ht!]
         \centering
         \includegraphics[width=1.3\textwidth,center]{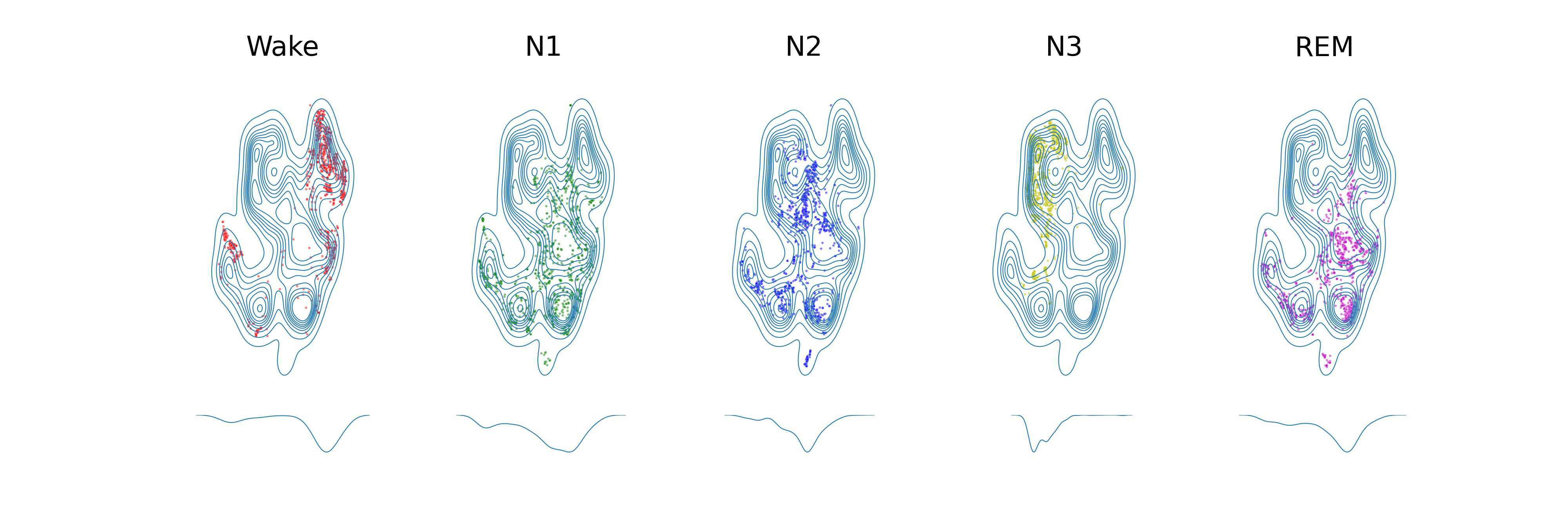}
         \caption{t-SNE visualization along with distribution plots on sleep-staging classes}
         \label{fig: tSNE}
     \hfill
\end{figure}

\par{\textbf{Sensitivity Analysis:}} We perform sensitivity analysis on sleepEDF dataset to study three hyperparameters ${\tau}_d$, $\lambda_1$, and $\lambda_2$ by changing a particular value while fixing the other two values. Fig.2(a) shows the results on ${\tau}_d$ varying from 0.1 to 10, the best results on MF1 are obtained when ${\tau}_d$ = 10. Fig.2(b) and Fig.2(c) show results on $\lambda_1$ and $\lambda_2$ respectively. We observe that the best results are obtained when setting $\lambda_1=1$ and $\lambda_2=1$ on MF1. The performance decreases when the values move away from 1 for both the hyperparameters. For $\lambda1$, decreasing its value hurts the performance severely compared to increasing it. But for $\lambda2$, increasing its value hurts the performance poorly compared to decreasing it.
\begin{figure}
     \centering
     \begin{subfigure}[b]{0.3\textwidth}
         \centering
         \includegraphics[width=\textwidth]{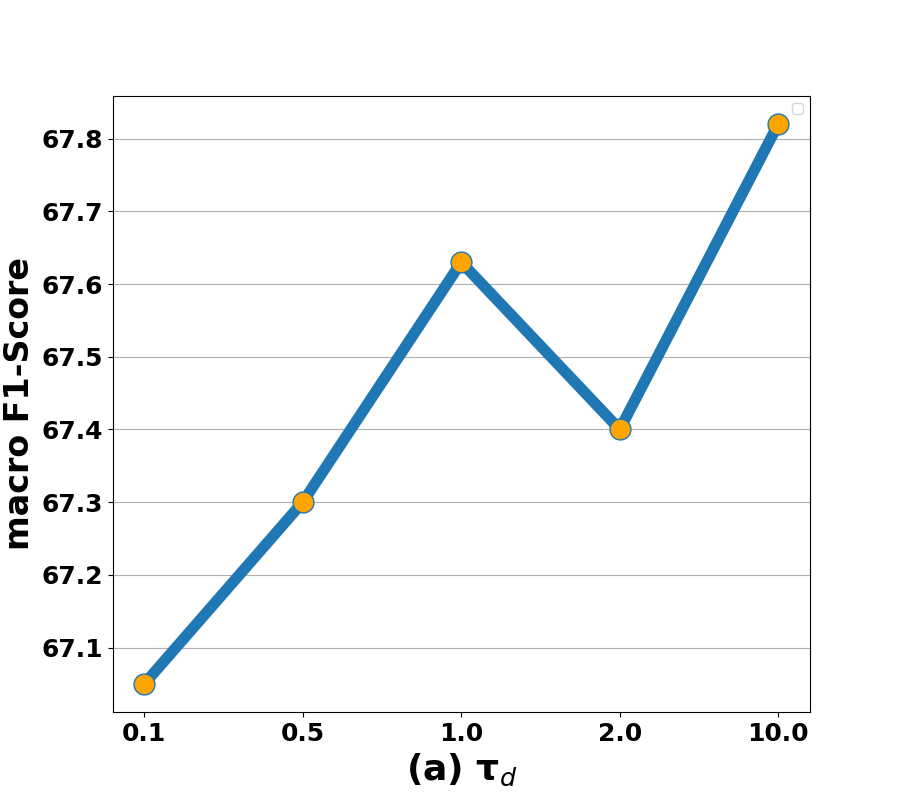}
         \label{fig: tau_d}
     \end{subfigure}
     \hfill
     \begin{subfigure}[b]{0.3\textwidth}
         \centering
         \includegraphics[width=\textwidth]{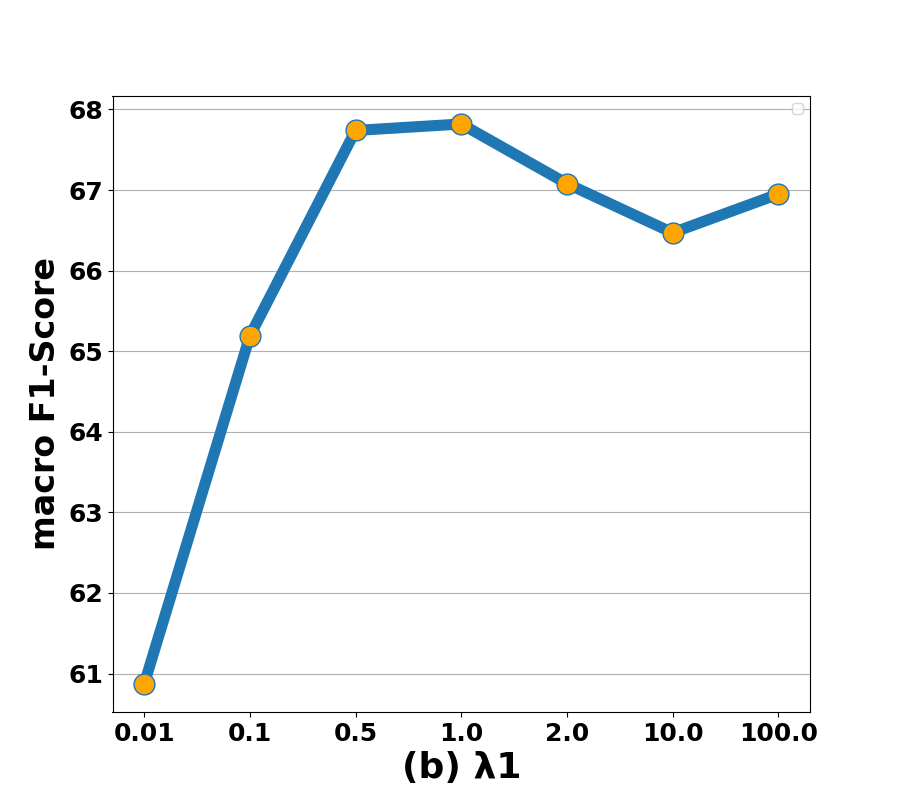}
         \label{fig: lambda1}
     \end{subfigure}
     \hfill
     \begin{subfigure}[b]{0.3\textwidth}
         \centering
         \includegraphics[width=\textwidth]{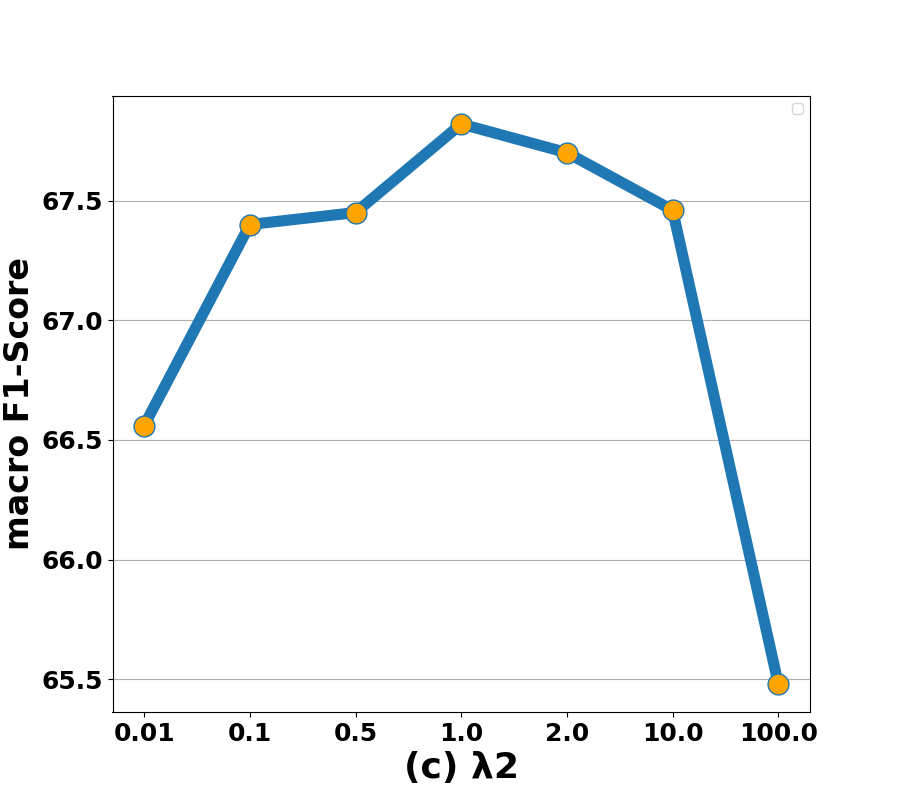}
         \label{fig: lambda2}
     \end{subfigure}     
        \caption{Sensitivity Analysis on three hyperparameters ${\tau}_d$, $\lambda_1$, and $\lambda_2$}
        \label{fig: sensitivity_analysis}
\end{figure}

\par{\textbf{Fine Tuning:}} We also present the finetuning results to observe the downstream task performance on sleep-staging, where a linear classifier along with the pre-trained encoder is trained completely on train data and evaluated on test data. The finetuning results on the transfer learning task is shown in Table 1. We observe that when the time-series encoder has completely finetuned, the difference between the baselines is marginal. But our model with diverse loss tends to beat other self-supervised and supervised baselines across all the metrics.

\begin{table}
\caption{Finetuning results on transfer learning tasks (SHHS $>$ sleepEDF)}

\centering
\begin{tabular}{|l|c c c|c c c|}
\hline
\multirow{1}{4em}{\textbf{Method}} 
& \textbf{Acc} & $\kappa$ & \textbf{MF1} \\
\hline
Single-View   &80.01&0.7142&71.55&\\
Simple Fusion &80.45&0.7212&71.64&\\
CMC &80.37&0.7203&71.77&\\
Ours    &80.20&0.7175&71.68&\\
Ours + diverse loss &\textbf{80.46}&\textbf{0.7213}&\textbf{71.88}&\\
\hline
Randomly Initialized &79.61&0.7082&70.18&\\
Supervised &80.18&0.7162&70.78&\\
\hline
\end{tabular}
\end{table}
\label{ft_results}

\par{\textbf{Ablation Study:}} We try to experiment effectiveness of the fusion contrastive module in our method on the sleepEDF dataset. On removing the fusion contrastive loss $L_{FF}$, a decrease in performance of 2.87\% (65.87) for MF1, 4.42\% (0.6547) for $\kappa$ and 2.47\% (76.13) for accuracy is observed, indicating that the fusion contrastive module is quite important in our setup.\\

\begin{figure}
     \centering
     \begin{subfigure}[b]{0.58\textwidth}
         \centering
         \includegraphics[width=\textwidth]{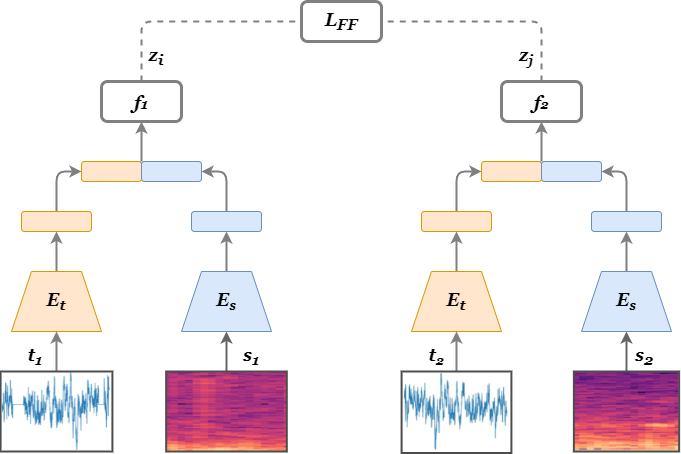}
         \caption{Simple Fusion}
         \label{fig: simple_fusion}
     \end{subfigure}
     \hfill
     \begin{subfigure}[b]{0.33\textwidth}
         \centering
         \includegraphics[width=\textwidth]{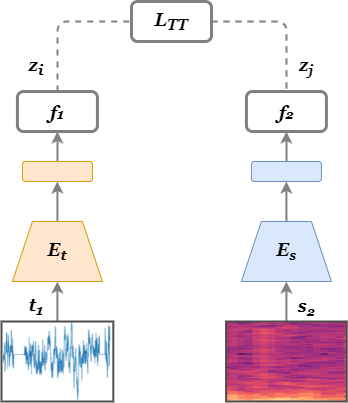}
         \caption{CMC}
         \label{fig: multi_view}
     \end{subfigure}
        \caption{Architecture overview of multi-view baselines Simple Fusion and CMC. (a) Augmentations $t_1$, $s_1$ are passed through $E_t$, $E_s$, and the concatenated outputs are passed through a projection head $f_1$ to obtain $\mathbf{z_i}$. Similarly, for augmented views  $t_2$, $s_2$, we extract $\mathbf{z_j}$. (b) Both the encoders are trained by sending $t_1$ to $E_t$ and $s_2$ to $E_s$, and the outputs from encoders are sent to their projection heads $f_1$, $f_2$ to obtain $\mathbf{z_i}$ and $\mathbf{z_j}$}
        \label{fig: model_architecture_2}
\end{figure}

\begin{figure}[ht!]
     \centering
     \begin{subfigure}[b]{0.3\textwidth}
         \centering
         \includegraphics[width=\textwidth]{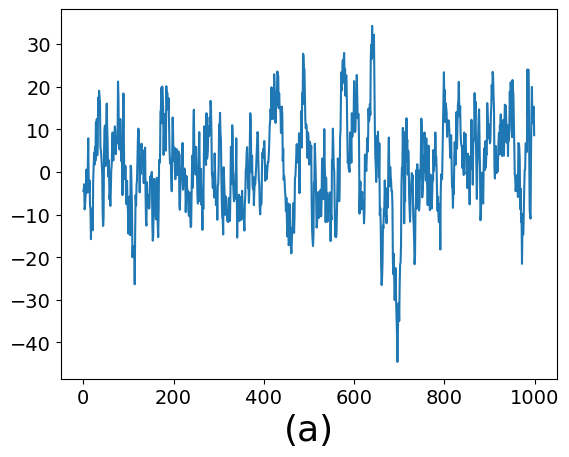}
         \label{fig: original_signal}
     \end{subfigure}
     \hfill
     \begin{subfigure}[b]{0.3\textwidth}
         \centering
         \includegraphics[width=\textwidth]{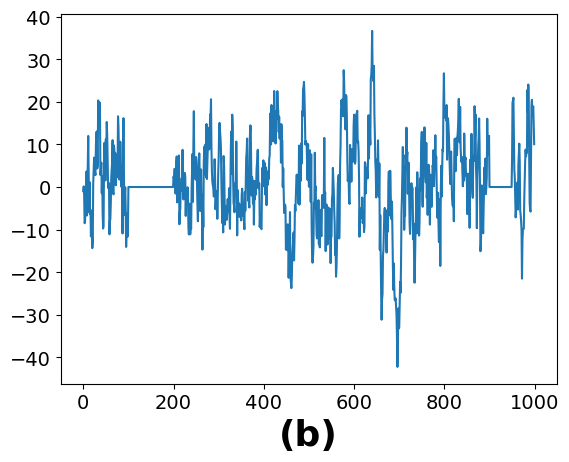}
         \label{fig: jitter_mask}
     \end{subfigure}
     \hfill
     \begin{subfigure}[b]{0.3\textwidth}
         \centering
         \includegraphics[width=\textwidth]{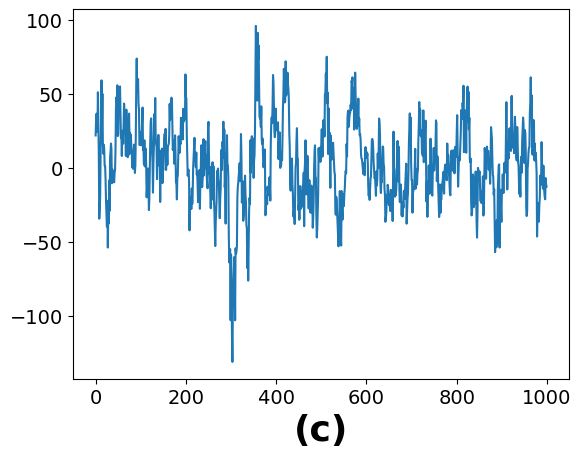}
         \label{fig: scaling_flip}
     \end{subfigure}     
        \caption{Different family of EEG augmentations used. (a) Original EEG signal. (b) Jittering, where random uniform noise is added to the EEG signal depending on its peak-to-peak values, along with masking, where signals are masked randomly. (c) Flipping, where the EEG signal is horizontally flipped randomly, and scaling, where EEG signal is scaled with Gaussian noise}
        \label{fig: augmentations}
\end{figure}

%
%
%

\bibliographystyle{splncs04}
\bibliography{References}
\end{document}